\documentclass[letterpaper, 10 pt, conference]{ieeeconf}
\IEEEoverridecommandlockouts
\usepackage{amsmath,amssymb,amsfonts}
\usepackage{graphicx}
\graphicspath{{./figures/}}
\usepackage[caption=false,font=footnotesize]{subfig}
\usepackage{booktabs}
\usepackage{multirow}
\usepackage{musicography}
\usepackage[hidelinks]{hyperref}

\title{\LARGE \bf
    Music-driven Robot Swarm Painting
}

\author{%
	Jingde Cheng and Gennaro Notomista%
	\thanks{The authors are with the Department of Electrical and Computer Engineering, University of Waterloo, Waterloo, ON, Canada {\tt\small j42cheng@uwaterloo.ca, gennaro.notomista@uwaterloo.ca}}%
}
    
\begin{document}

\maketitle
\thispagestyle{empty}
\pagestyle{empty}

\begin{abstract}
This paper proposes a novel control framework for robotic swarms capable of turning a musical input into a painting. The approach connects the two artistic domains, music and painting, leveraging their respective connections to fundamental emotions. The robotic units of the swarm are controlled in a coordinated fashion using a heterogeneous coverage policy to control the motion of the robots which continuously release traces of color in the environment. The results of extensive simulations performed starting from different musical inputs and with different color equipments are reported. Finally, the proposed framework has been implemented on real robots equipped with LED lights and capable of light-painting.
\end{abstract}

\section{Introduction}

Robots have long been used for automation and monitoring, yet their roles in creative domains, such as painting and music, are comparatively new. Most robot painting research focuses on robot arm painting, where manipulators apply brushstrokes to a canvas. In contrast, swarm painting studies (e.g. \cite{santos2020interactive}) have shown that multiple mobile robots can collectively create art by depositing paint or leaving colored trails. These approaches often rely on predefined, static inputs (e.g., user-defined color-density maps) and do not consider dynamic data sources. Separately, research at the intersection of robots and music has explored choreographed movements, music performance, and emotional expression through motion. For instance, swarms have been programmed to move in sync with musical tempos or to display emotions via their trajectories. However, to date there has been little investigation into how music might directly influence a swarm's creation of visual art.

Both robotics researchers and artists have moved towards each other in trying to understand the benefits that robotics may bring to art and vice versa. Examples of artistic exhibitions where robots play a central role are numerous, ranging from dancing robots \cite{bi2018real} to humanoid robots improvising on jazz standards alongside humans \cite{hoffman2010gesture}, from robot-aided painting \cite{santos2021motions,chen2022gtgraffiti} to flying robots performing on stage \cite{ackerman2014flying}. Robots have also been the subject of art exhibitions. In \cite{dean2008robotic} the authors conceived a robotic chair, able to locomote, which took part in an exhibition at the Vancouver Contemporary Art Gallery. The concept of robotics culture is introduced in \cite{dunstan2016cultural}, where the authors propose a robotic entity that participates in the development of material and non-material culture. In \cite{jochum2016cultivating} the Freudian and Morian concepts of the Uncanny and their influence on artists working with robots is analyzed, with a special focus on anthropomorphic phenomena and centered around three specific artworks. Another prominent example of the use of robots as forms of art is the one presented in \cite{vlachos2018heat}, where the authors analyze the human-exoskeleton interaction based on methods from the performing arts, art-based rehabilitation therapy, and social robotics.

\begin{figure}
    \centering
    \scalebox{-1}[1]{\includegraphics[trim={2cm 1cm 2cm 6cm},clip,width=0.9\linewidth]{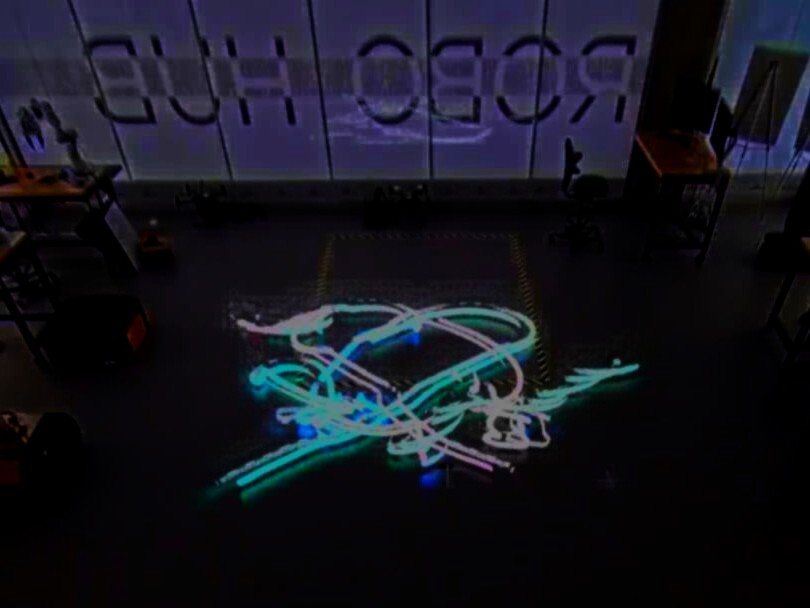}}
    \caption{Six DJI Robomaster EP robots light-painting the ``lost'' Waltz in A minor---attributed to Frédéric Chopin and unearthed on October 27, 2024.}
    \label{fig:lost_waltz}
\end{figure}

While the direction from art to robotics is well represented in the literature, the influence of art in robotics has received considerable less attention. Questions like \textit{``Can we leverage methodologies coming from different forms of art to control the behavior of robotic systems?''} and \textit{``How can a form of art be connected to another one thanks to the aid of intelligent robots?''} have not been considered yet. This paper aims to close this gap by introducing a music-driven swarm painting system. Our approach consists in extracting features from music---such as chord progressions and tempo---and mapping them to color-density functions for a team of mobile painting robots. By integrating music analysis with coverage control, our system continuously adapts the robots' movements, thus allowing them to \textit{paint music} in real time (see Fig.~\ref{fig:lost_waltz}). To the best of our knowledge, no prior work has combined swarm robotics, coverage control, and music-to-color mapping in this way.

The focus of robot painting has been mainly on the use of robotic arms which are capable of reproducing input images \cite{scalera2019watercolour}. The subject of such images vary from portraits to still lives to abstract paintings \cite{schubert2017optimal}. The use of robotic swarms to generate paintings is mainly confined to computer graphics technologies \cite{urbano2005playing}, which are based on a human input that needs to be followed as faithfully as possible. In other works, such as \cite{moura2016machines} and \cite{santos2020interactive}, teams of robots follow specific behaviors and their actions are completely autonomous, making the interaction between human painters and robotic swarms non-existent. Recent works have considered the co-existence of human painters alongside robots \cite{chung2018online}, however the interaction between them is limited as the behavior of the robots is not influenced by the actions of the human painter. The lack of more complex and expressive interaction modalities between humans and robots is one of the main motivations for the approach presented in this paper.

The research landscape concerned with the interplay between music and robotics is quite different from the case of painting and robotics. In fact, while there have been attempts of making robots be able to reproduce musical compositions by playing an instrument (e.g. \cite{kuwabara2006development,shibuya2007toward}, just to name a few) in many cases is the music that influences the behavior of the robots. A notable example is the humanoid robot developed to improvise alongside human musicians that is able to respond to stimuli coming from the music played by other musicians \cite{hoffman2010gesture}. In this application, the human musicians are also influenced by the robot playing in a collaborative environment typical of jazz music. Similar approaches can be found in \cite{mizumoto2008robot}.

The contributions of this paper with respect to the state of the art are twofold. First, swarms of robotic units which are not anthropomorphic are considered. By preventing to directly transferring human-like features to robots, this constraint results in novel robot control approaches suitable for artistic applications. Moreover, this paper aims to understand not only the interplay between robotics and a single form of art, but rather how the connections between different forms of art can suggest novel interaction modalities between robots and humans. These contributions advance the technical and artistic aspects of robotic art and, in addition, they suggest broader social and economic impact of robotics. In particular: (i) In museums, music-driven robot swarms are envisioned to adaptively respond to ambient sound and user interaction, creating dynamic visual feedback that makes the environment more engaging; (ii) As public art installations, they could help bringing people together and making creative experiences more accessible; (iii) Economically, this approach opens new avenues for innovative industrial design; (iv) Last, but not least, in education, the developed system can serve as a hands-on tool to teach students about robotics, music, and emotions in an interactive fashion.

\section{Proposed Approach}

\subsection{Problem Statement}

\begin{figure}
	\centering
	\includegraphics[width=0.48\textwidth]{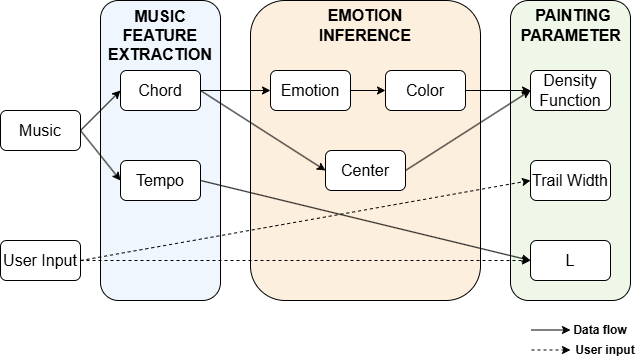}
	\caption{Music-to-painting pipeline for the robot swarm. First, a \texttt{Music} input is analyzed to obtain \texttt{Chords} and \texttt{Tempo}. Then, \texttt{Chords} are mapped to \texttt{Emotions}, which determine both the \texttt{Color} and the goal positions for the robots (\texttt{Center}), together forming a \texttt{Density Function}. The \texttt{Tempo} together with the \texttt{User Input}, sets a motion parameter of the robot control algorithm (\texttt{L}). The user may also set the color \texttt{Trail Width}. The resulting color \texttt{Density Function}, \texttt{Trail Width}, and \texttt{L} are finally used to control the robot swarm.}
	\label{fig:figure1}
\end{figure}

In this paper we develop a system that enables a swarm of painting robots to autonomously translate music into visual art.  The core problem lies in creating a framework that can process musical features, such as chords and tempos, interpret their emotional content, and translate these elements into appropriate color densities for robotic painting. Our approach also requires integrating two techniques from different domains: coverage control algorithms for robotic swarms and music feature extraction and analysis. The central challenge is designing a cohesive system that not only processes musical input but also coordinates multiple painting robots to create adaptive, expressive art that reflects the emotional essence of the music being played. The overall approach is depicted in the block diagram in Fig.~\ref{fig:figure1} and will be explained in the following sections.

\subsection{Coverage Control for Painting}

Coverage control is a coordinated control algorithm that is used to optimally distribute a team of robots across a domain---in this case the \textit{canvas}---by leveraging density functions that represent desired distributions \cite{cortes2004coverage}. Traditionally used for distributed sensing and monitoring, coverage control ensures that the position of each robot minimizes a cost defined by a density function, effectively guiding it to regions of greater importance.

In our robotic painting system, a Gaussian density function is defined as $\phi_{\beta}(q) = \frac{K}{2 \pi \sigma_x \sigma_y}
e^{- \frac{1}{2} \left( \left(\frac{q_x - \mu_x}{\sigma_x}\right)^2 + \left(\frac{q_y - \mu_y}{\sigma_y}\right)^2 \right)}$, where \(q = [q_x, q_y]\in\mathbb R^2\) represents a point on the canvas, \(\mu_x, \mu_y\) define the center of the color density, \(\sigma_x, \sigma_y\) control its spread, and \(K\) determines the color intensity at the center. To handle heterogeneous resources---representing robots carry different subsets of colors---the system integrates multiple density functions into a unified control strategy \cite{santos2020interactive}. For each robot \(i\) and color \(j\), we compute the heterogeneous mass $M^j_i(x) = \int_{V^j_i(x)} \phi_j(q) \, dq$
which measures the total weight or importance of the density function \(\phi_j(q)\) within the robot's Voronoi cell \(V^j_i(x)\), and the center of mass $C^j_i(x) =
\left. \int_{V^j_i(x)} q \, \phi_j(q) \, dq \middle/ \int_{V^j_i(x)} \phi_j(q) \, dq, \right.$, which represents the weighted average position of the density in that region, indicating the optimal target position for the robot. Using a gradient descent approach, the control law directs each robot towards the center of mass of its region: $u_i = \sum_{j \in P(i)} M^j_i(x) \bigl(C^j_i(x) - x_i \bigr)$, where \(x_i\) is the current position of robot \(i\) and \(P(i)\) represents the set of colors or paint resources available to it.

Finally, the system applies CMY (cyan, magenta, yellow) color mixing model to merge individual color density functions and produce a wide spectrum of colors from minimal primary pigments. In particular, each robot determines the proportion of pigments denoted by $\alpha^j_i = \left. M^j_i(x) \middle/ \sum_{k \in P(i)} M^k_i(x)\right.$, where $\alpha^j_i \in [0,1]$ and \(M^j_i(x)\) is the heterogeneous mass of color \(j\) in robot \(i\)'s region of dominance, and \(P(i)\) is the set of pigments available to robot \(i\). This formulation ensures that the pigment used by the robots reflects the color demands, leading to an accurate representation of the desired color map on the canvas.

In summary, by controlling the swarm using density functions that encode the color and spatial distributions, our system enables robots to dynamically adapt their position and pigment usage. This ensures that the final painting accurately reflects the intended visual design while effectively managing the heterogeneous capabilities of the robotic team.

\subsection{Tempo-controlled $L$}

In our differential drive model for the robot motion, the parameter $L$ represents the inverse scaling factor for angular velocity. We modify the traditional approach by using $1/L$ as the angular velocity scalar, creating an inverse relationship where smaller $L$ values produce higher turn rates and larger values yield lower turn rates. To dynamically adjust this parameter based on music, we implement a linear mapping function that correlates $L$ with tempo $t$ as $L = L_{max} - (L_{max} - L_{min}) \min(t, t_{max}) / t_{max}$, with $0<L_{min}<L_{max}$ and $t_{max}>0$.
This ensures that faster tempos lead to lower $L$, resulting in more rapid turning responses, while slower tempos increase $L$ for more gradual turns.

\begin{figure}
	\centering
	\includegraphics[trim={0 1.5cm 0 2.3cm},clip,width=0.6\linewidth]{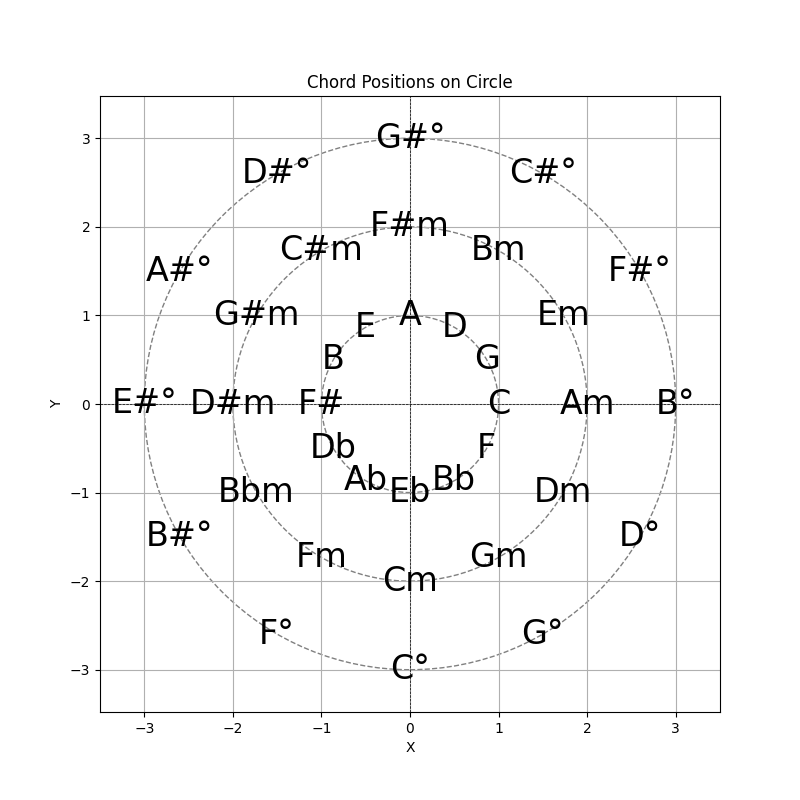}
	\caption{Chord wheel: a mapping from chords to canvas locations.}
	\label{fig:figure3}
\end{figure}

\subsection{Chord-to-position Mapping}
\label{sec:center_mapping}

To spatially represent chord progressions, we employ a \textit{chord wheel}, a circular diagram that arranges chords according to their harmonic relationships (see Fig.~\ref{fig:figure3}). Adjacent chords on the wheel share strong harmonic similarity, while chords lying farther apart indicate greater tension. By mapping each detected chord to a coordinate on this wheel, we create a dynamic spatial reference for how color should be placed on the canvas. When chords change, the system adjusts the center of the corresponding color density function, shifting painting activity to new regions. This ensures a continuous, real-time translation of harmonic motion into spatial variations on the canvas, thus effectively realizing the expressive connection between the music and the visual outcome. Once the chord-to-color and chord-to-position mappings are established, the system generates color density functions that reflect the chord wheel's spatial arrangement.

\subsection{User Interaction}
In our proposed approach, a human painter retains creative control over the swarm painting process through an intuitive graphical interface. By clicking on the simulator canvas, users can specify the center of the color density functions, effectively guiding where the painting will occur. In addition, an adjustable slider allows for real-time manipulation of robot motion control parameters. This interactive design ensures that the automated coverage control of the swarm remains responsive to the artistic vision of the user, striking a balance between the autonomous behavior of the system and human creativity.

\section{Simulations and Experiments}

\begin{table}
\centering
\caption{Music pieces used in simulations}
\begin{tabular}{p{0.22\linewidth}|p{0.68\linewidth}}
\toprule
\textbf{Author} & \textbf{Piece} \\
\midrule
\textbf{Bach} & Prelude No. 1 in C Major (BWV 846) \\
& Prelude No. 2 in C Minor (BWV 847) \\
& Prelude No. 6 in D Minor (BWV 875) \\
\hline
\textbf{Mozart} & Piano Sonata No. 11 in A Major (K. 331) \\
& Piano Sonata No. 16 in C Major (K. 545) \\
& Piano Sonata No. 12 in F Major (K. 332) \\
\hline
\textbf{Beethoven} & Piano Sonata No. 14 in C\# Minor \\
& Piano Sonata No. 8 in C Minor \\
& Piano Sonata No. 1 In F Minor \\
\hline
\textbf{Chopin} & Waltz in D\fl~Major (Op. 64, No. 1) \\
& Waltz in C\sh~Minor (Op. 64, No. 2) \\
& Waltz in A\fl~Major (Op. 69, No. 1) \\
\hline
\textbf{Satie} & Gymnopédie No. 1 \\
& Gymnopédie No. 2 \\
& Gymnopédie No. 3 \\
\bottomrule
\end{tabular}
\label{table:musicpieces}
\end{table}

\begin{table}
	\centering
	\caption{Mapping between chords and emotions}
	\begin{tabular}{p{0.385\linewidth}|p{0.52\linewidth}}
		\toprule
		\textbf{Chord} & \textbf{ Emotions}\\
		\midrule
		Major tonic & Serenity, Acceptance, Trust \\ \hline
		Minor tonic & Grief, Sadness, Anger \\ \hline
		Natural minor & Vigilance, Aggressiveness \\ \hline
		Dominant & Joy, Ecstasy, Amazement \\ \hline
		Seventh & Rage, Grief, Disgust \\ \hline
		Secondary dominant & Surprise, Bittersweet joy \\ \hline
		Major subdominant & Joy, Admiration, Serenity \\ \hline
		Major subdominant 7th & Pensiveness, Sadness, Yearning \\ \hline
		Added sixth in a major & Love, Trust, Acceptance \\ \hline
		Added sixth in a minor & Grief, Sadness, Remorse \\ \hline
		Neapolitan sixth & Grief, Sadness, Pensiveness \\ \hline
		Diminished seventh & Fear, Despair, Terror \\ \hline
		Augmented & Amazement, Surprise, Ecstasy \\ \hline
		Minor sixth & Fear, Anxiety, Apprehension \\ \hline
	\end{tabular}
	\label{table:chord_emotion_mapping}
	\vspace{-0.3cm}
\end{table}

Several music inputs, varying in genre and tempo, have been tested to examine their impact on the behavior of the swarm. The list of compositions is reported in Table~\ref{table:musicpieces}.

\subsection{Mapping Music to Color Density Functions}

To enable a swarm of painting robots to autonomously generate visual art from musical input through coverage control, we propose a system that combines music analysis, emotion modeling, and spatial mapping into color density functions. Figure~\ref{fig:figure1} illustrates the data flow, and our approach proceeds through the following key stages.

\subsubsection{Music Feature Extraction}

We begin by analyzing an audio stream to extract two primary musical features, chords and tempo. These features form the basis for mapping music to emotions, ultimately resulting in visual representations.
\underline{Chords:} The system detects chord progressions, recognizing that each chord evokes distinct emotional associations. These relations are summarized in Table~\ref{table:chord_emotion_mapping}. For instance, a major chord typically suggests feelings of happiness or stability, while a minor chord is related to sadness or pensiveness. \underline{Tempo:} The music speed influences both the intensity and breadth of robotic movements on the canvas. A fast tempo encourages broader, more dynamic strokes, whereas a slower tempo promotes smoother and wider painting paths.

\subsubsection{Mapping Chords to Color}
\label{sec:chord_to_color}

Once chords are identified, leveraging theory of musical equilibration \cite{willimek2013music}, we map each chord to an emotional state. These emotion labels are subsequently mapped to color codes derived from Plutchik's wheel of emotions \cite{plutchik2001integration}, providing a structured way to select and blend hues that represent each emotional state.

\subsection{Simulations}

Extensive simulations have been run to study the effect of all the system parameters. The trail width $w \in \{ 10, 15, 20 \}$ units was adjusted to assess its influence on the resulting artwork. The control parameter $L \in \{1, 3, 5\}$, determining the responsiveness of the robots, trades off between agile maneuvering and smooth motion. To evaluate the effect of swarm size on collective behavior and coordination overhead, simulations were performed with varying numbers of robots $N \in \{ 6, 9, 12 \}$. In addition, differentiation in color equipment was explored by equipping robots with 1 to 3 distinct colors equipment, either uniformly or in mixed configurations. Finally, a real-time human-robot interaction component allowed users to dynamically adjust $w$, $L$, and density center during the simulations, providing insight into the system responsiveness under user intervention. 

\begin{figure}
	\vspace{-0.35cm}
	\centering
	% bach
	\subfloat[Prelude and Fugue in C (WTK, Book I, No. 1), BWV 846]{\includegraphics[trim={3cm 2.2cm 3.8cm 1.3cm},clip,width=0.33\linewidth]{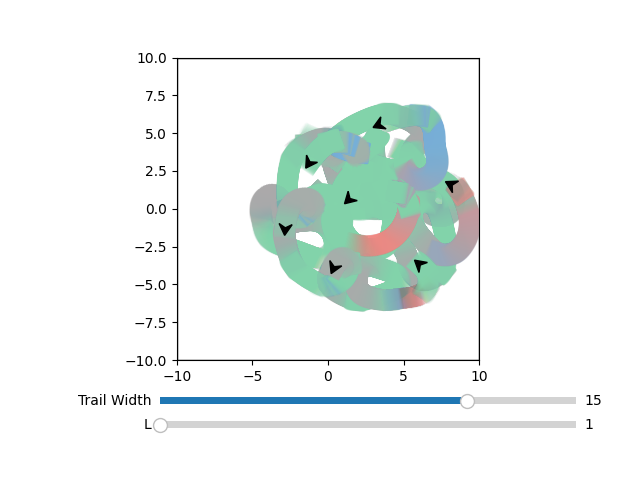}}\hfill
	\subfloat[Prelude and Fugue in C minor (WTK, Book I, No. 2), BWV 847]{\includegraphics[trim={3cm 2.2cm 3.8cm 1.3cm},clip,width=0.33\linewidth]{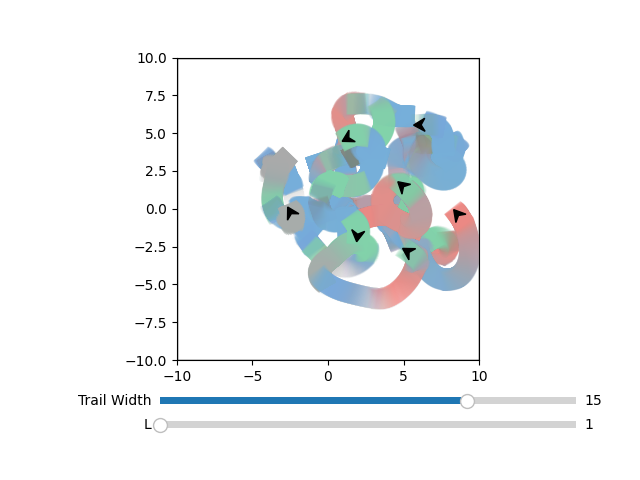}}\hfill
	\subfloat[Prelude and Fugue in D minor (WTK, Book II, No. 6), BWV 875]{\includegraphics[trim={3cm 2.2cm 3.8cm 1.3cm},clip,width=0.33\linewidth]{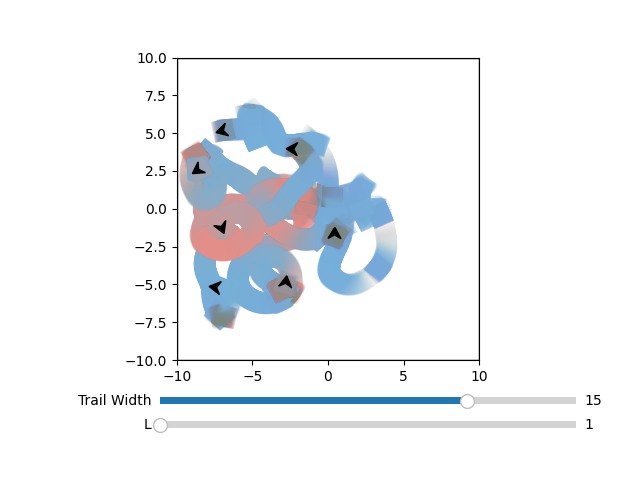}}\\[-0.2cm]
	% mozart
	\subfloat[Piano Sonata No. 12 in F, K. 332, 2. Adagio]{\includegraphics[trim={3cm 2.2cm 3.8cm 1.3cm},clip,width=0.33\linewidth]{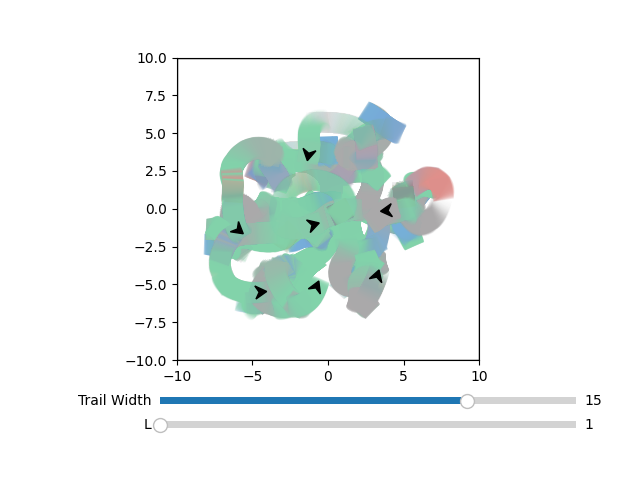}}\hfill
	\subfloat[Piano Sonata No. 11 in A Major, K. 331, 3. Alla Turca. Allegretto]{\includegraphics[trim={3cm 2.2cm 3.8cm 1.3cm},clip,width=0.33\linewidth]{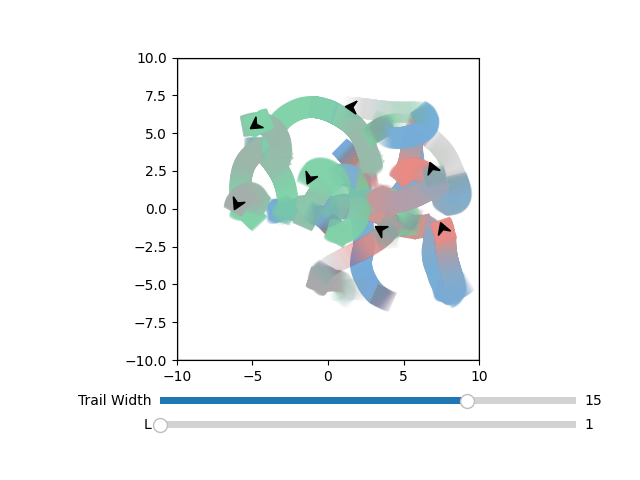}}\hfill
	\subfloat[Piano Sonata No. 16 in C, K. 545, Sonata facile, 1. Allegro]{\includegraphics[trim={3cm 2.2cm 3.8cm 1.3cm},clip,width=0.33\linewidth]{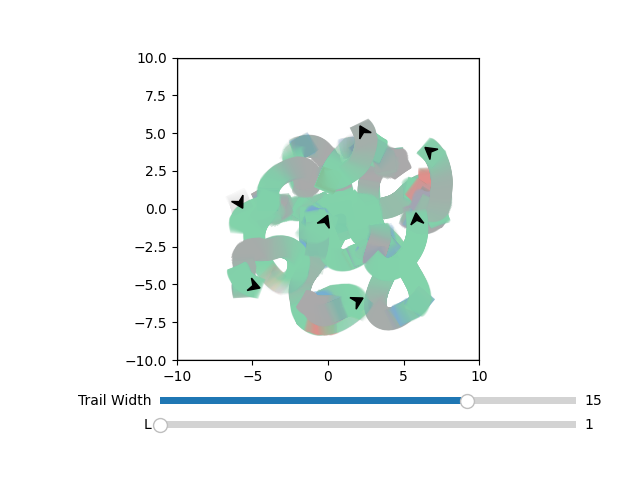}}\\[-0.2cm]
	% beethoven
	\subfloat[Piano Sonata No. 14 in C\sh Minor, Op. 27/2, I. Adagio sostenuto]{\includegraphics[trim={3cm 2.2cm 3.8cm 1.3cm},clip,width=0.33\linewidth]{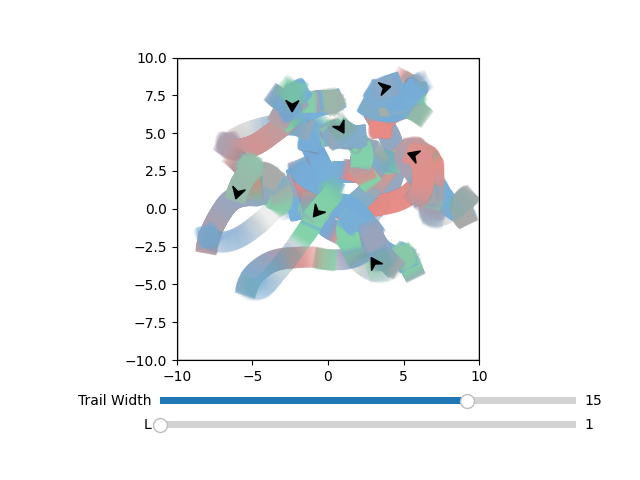}}\hfill
	\subfloat[Piano Sonata No. 8 in C Minor, Op. 13, II. Adagio cantabile]{\includegraphics[trim={3cm 2.2cm 3.8cm 1.3cm},clip,width=0.33\linewidth]{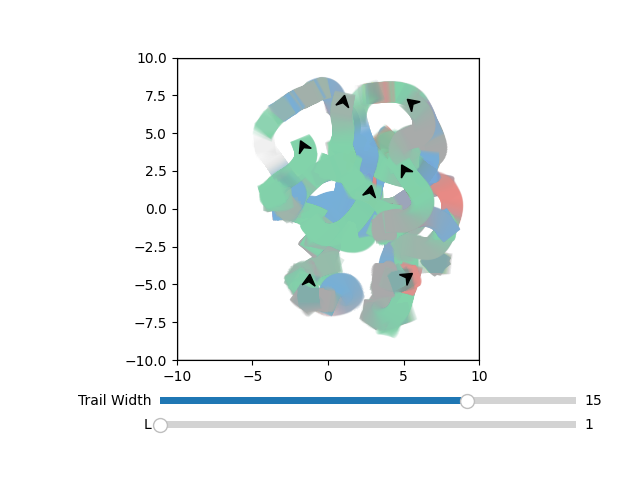}}\hfill
	\subfloat[Piano Sonata No. 1 in F Minor, Op. 2, No. 1, 1. Allegro]{\includegraphics[trim={3cm 2.2cm 3.8cm 1.3cm},clip,width=0.33\linewidth]{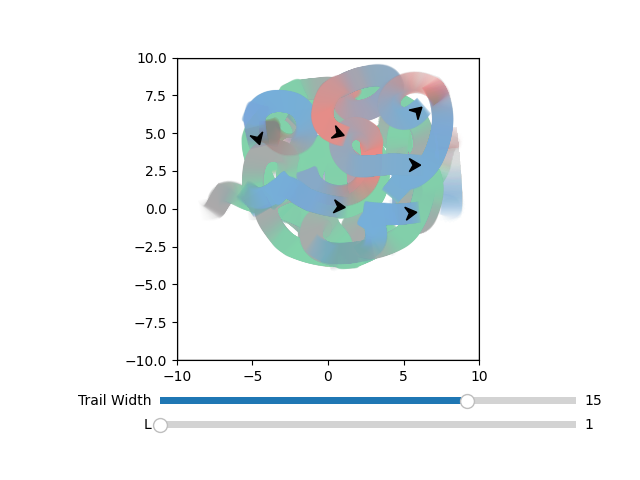}}\\[-0.2cm]
	% chopin
	\subfloat[2 Waltzes, Op. posth. 69, No. 1 in A\fl Major]{\includegraphics[trim={3cm 2.2cm 3.8cm 1.3cm},clip,width=0.33\linewidth]{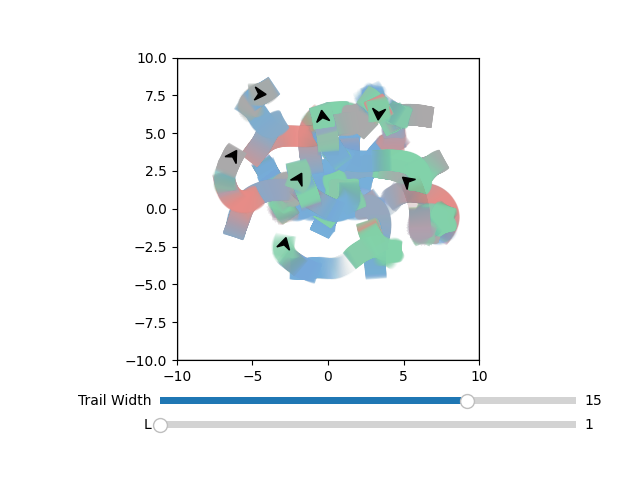}}\hfill
	\subfloat[3 Waltzes, Op. 64, No. 1 in D\fl Major]{\includegraphics[trim={3cm 2.2cm 3.8cm 1.3cm},clip,width=0.33\linewidth]{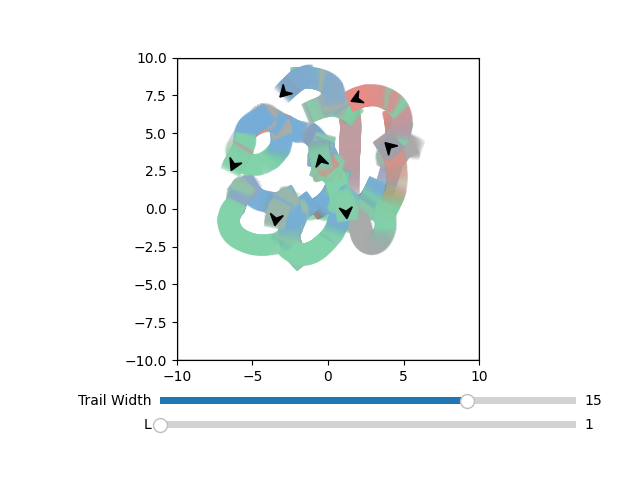}}\hfill
	\subfloat[3 Waltzes, Op. 64, No. 2 in C\sh Minor]{\includegraphics[trim={3cm 2.2cm 3.8cm 1.3cm},clip,width=0.33\linewidth]{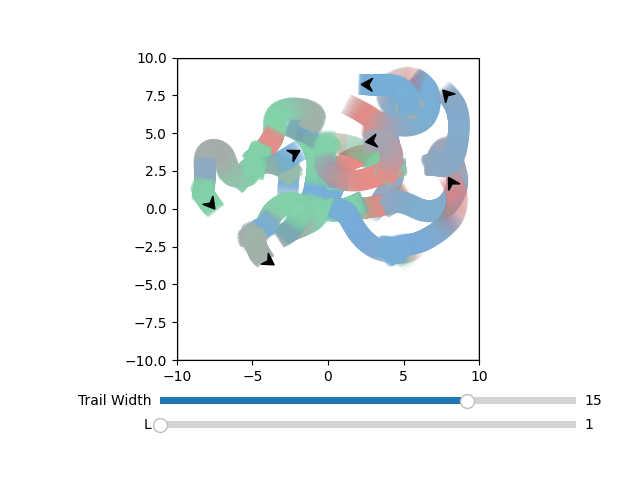}}\\[-0.2cm]
	% satie
	\subfloat[Gymnopédie No. 1]{\includegraphics[trim={3cm 2.2cm 3.8cm 1.3cm},clip,width=0.33\linewidth]{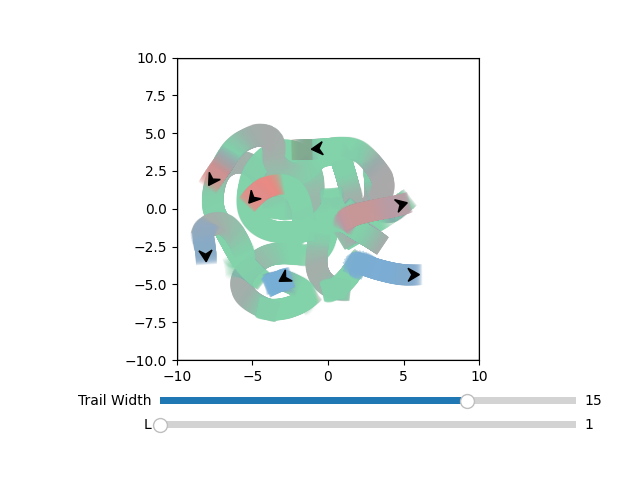}}\hfill
	\subfloat[Gymnopédie No. 2]{\includegraphics[trim={3cm 2.2cm 3.8cm 1.3cm},clip,width=0.33\linewidth]{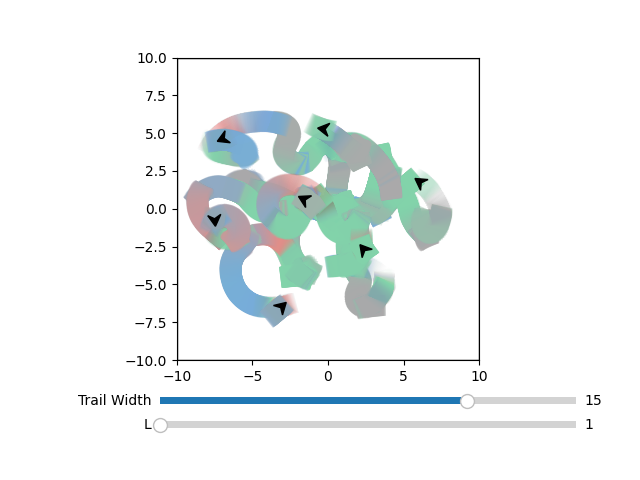}}\hfill
	\subfloat[Gymnopédie No. 3]{\includegraphics[trim={3cm 2.2cm 3.8cm 1.3cm},clip,width=0.33\linewidth]{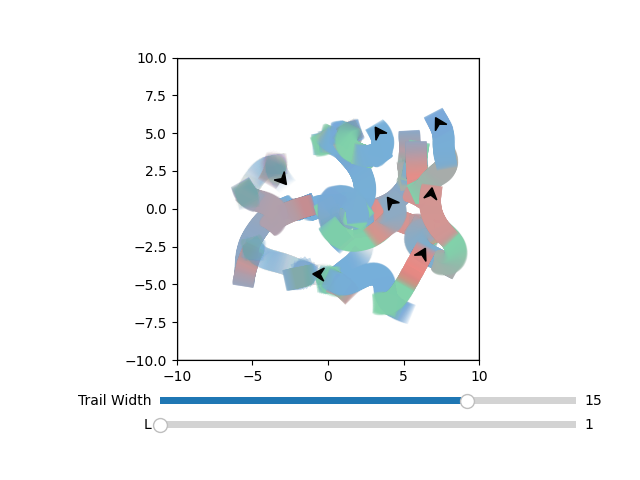}}
	\caption{Paintings resulting from different music pieces.}
	\label{fig:grid_pieces}
	\vspace{-0.5cm}
\end{figure}

\begin{figure}
	\centering
	\subfloat[Setup 1]{\includegraphics[trim={3cm 2.2cm 3.8cm 1.3cm},clip,width=0.33\linewidth]{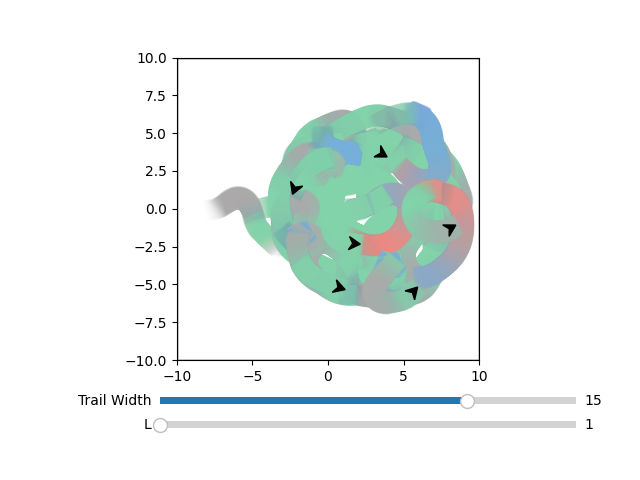}}\hfill
	\subfloat[Setup 2]{\includegraphics[trim={3cm 2.2cm 3.8cm 1.3cm},clip,width=0.33\linewidth]{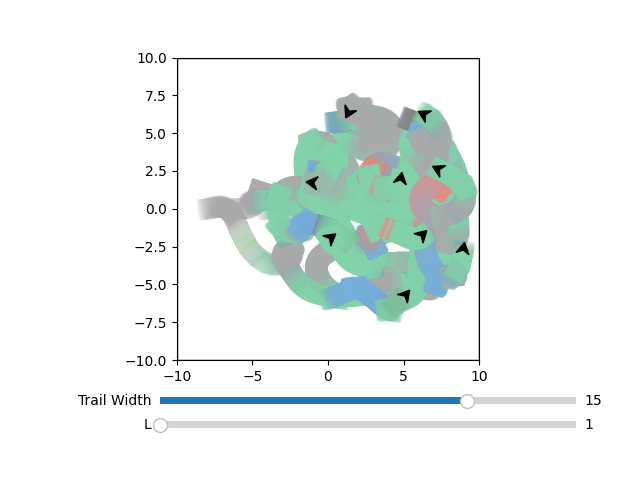}}\hfill
	\subfloat[Setup 3]{\includegraphics[trim={3cm 2.2cm 3.8cm 1.3cm},clip,width=0.33\linewidth]{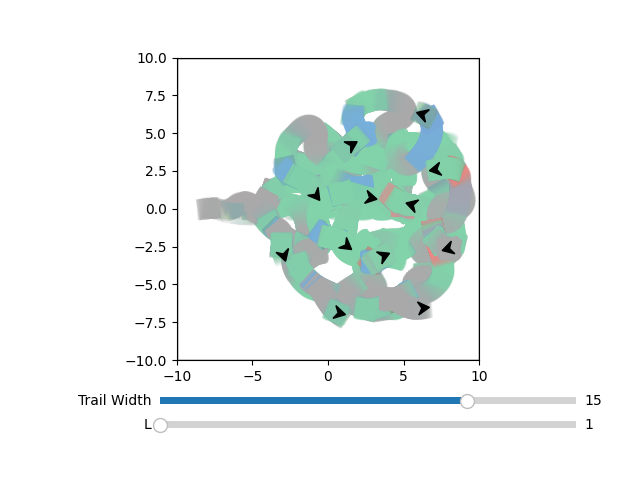}}\\[-0.2cm]
	\subfloat[Setup 4]{\includegraphics[trim={3cm 2.2cm 3.8cm 1.3cm},clip,width=0.33\linewidth]{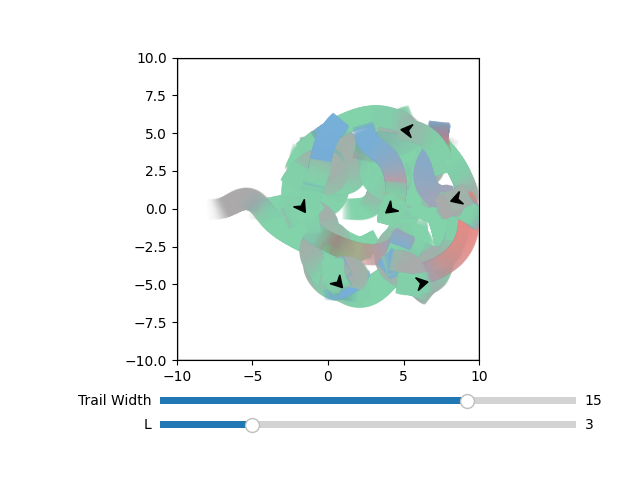}}\hfill
	\subfloat[Setup 5]{\includegraphics[trim={3cm 2.2cm 3.8cm 1.3cm},clip,width=0.33\linewidth]{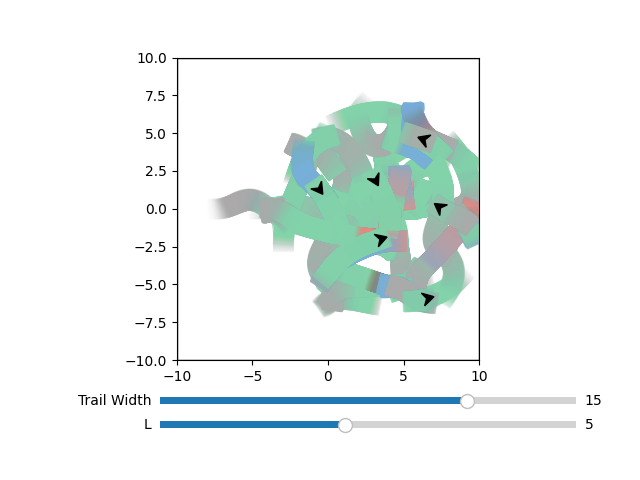}}\hfill
	\subfloat[Setup 6]{\includegraphics[trim={3cm 2.2cm 3.8cm 1.3cm},clip,width=0.33\linewidth]{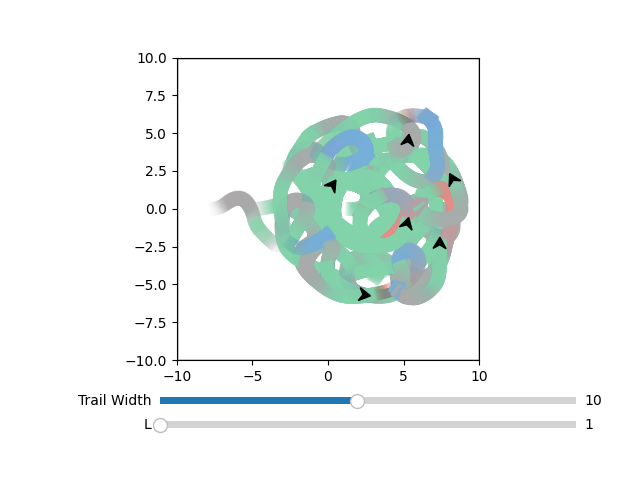}}\\[-0.2cm]
	\subfloat[Setup 7]{\includegraphics[trim={3cm 2.2cm 3.8cm 1.3cm},clip,width=0.33\linewidth]{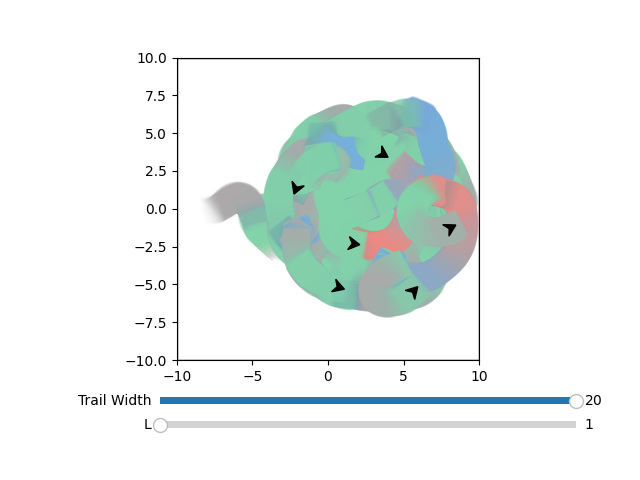}}\hfill
	\subfloat[Setup 8]{\includegraphics[trim={3cm 2.2cm 3.8cm 1.3cm},clip,width=0.33\linewidth]{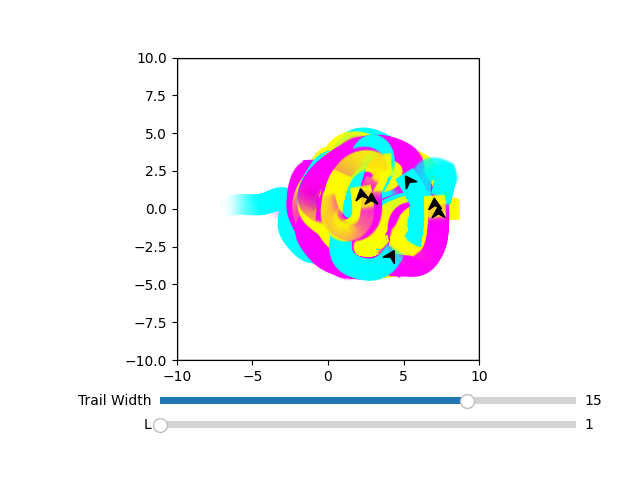}}\hfill
	\subfloat[Setup 9]{\includegraphics[trim={3cm 2.2cm 3.8cm 1.3cm},clip,width=0.33\linewidth]{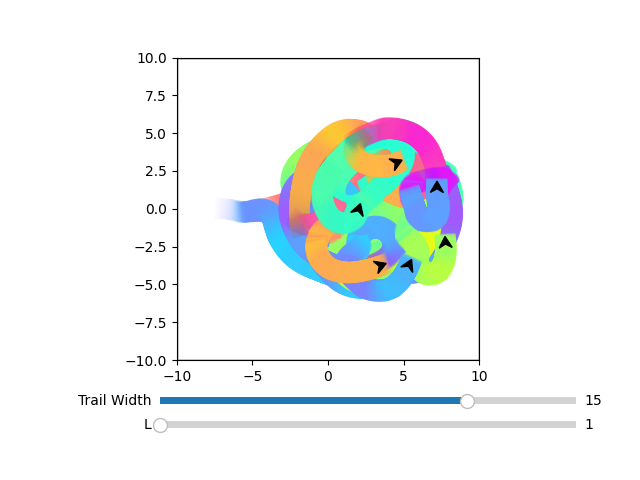}}\\[-0.2cm]
	\subfloat[Setup 10]{\includegraphics[trim={3cm 2.2cm 3.8cm 1.3cm},clip,width=0.33\linewidth]{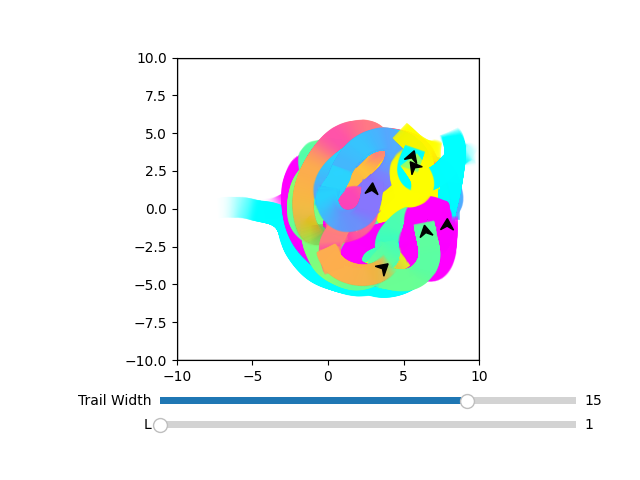}}\hfill
	\subfloat[Setup 11]{\includegraphics[trim={3cm 2.2cm 3.8cm 1.3cm},clip,width=0.33\linewidth]{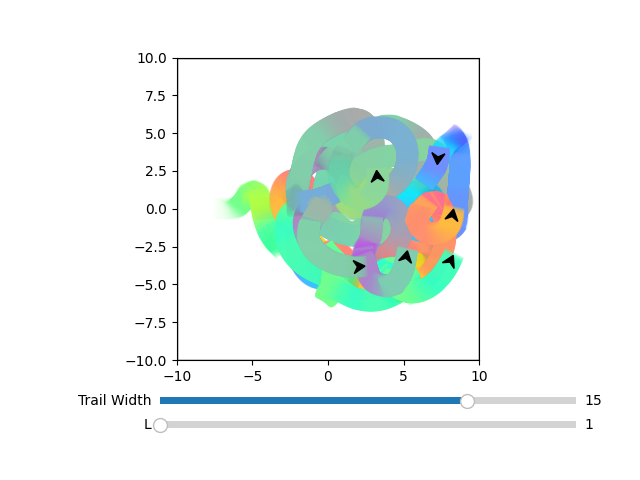}}\hfill
	\subfloat[Setup 12]{\includegraphics[trim={3cm 2.2cm 3.8cm 1.3cm},clip,width=0.33\linewidth]{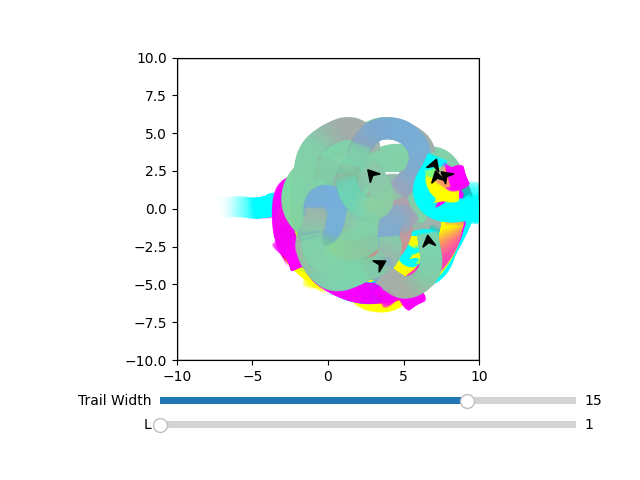}}\\[0.1cm]
	\subfloat[Setup 13]{\includegraphics[trim={3cm 2.2cm 3.8cm 1.3cm},clip,width=0.33\linewidth]{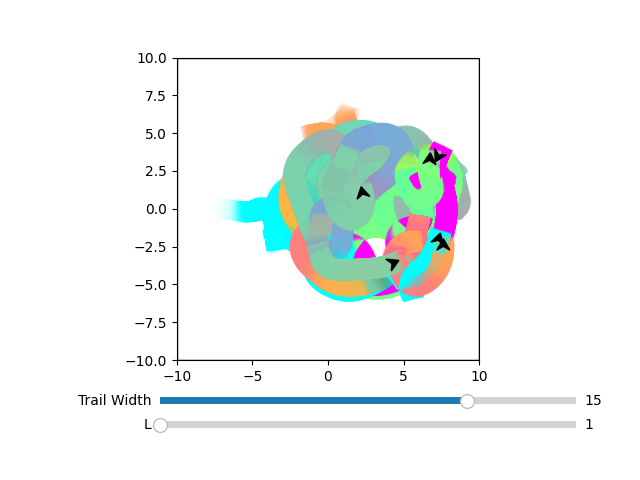}}
	\caption{Paintings obtained with the different setups listed in Table~\ref{table:setup}.}
	\label{fig:grid_setup}
\end{figure}

The simulation results are illustrated in Figures~\ref{fig:grid_pieces}, \ref{fig:grid_setup}, and \ref{fig:grid_setup2} and highlight the effect of the system parameters on the generated robotic artwork. With reference to Table~\ref{table:setup}, \underline{Swarm Size ($N$)}: Increasing the number of robots from 6 to 12 (Setups 1--3) led to more intricate and denser patterns, as more agents coordinate to optimally cover the color densities based on their color equipments; \underline{Control Parameter ($L$)}: The variation of $L$ (Setups 4--6) demonstrated a clear trade-off between agility and smoothness, where lower values of $L$ (e.g., Setup 4 with $L = 3$) produced more responsive movements, while higher values (e.g., Setup 5 with $L = 5$) resulted in smoother and more gentle trails. This suggests that lower $L$ improves sensitivity to external inputs (e.g., music dynamics), while higher $L$ leads to slower and rounder robots' trajectories; \underline{Color Equipment Differentiation}: Setups 8--13, which varied the number and distribution of color equipment, illustrated how visual contrast can be obtained in swarm-generated art. When robots were equipped with multiple colors in mixed configurations (e.g., Setup 9), the resulting patterns showed greater diversity in tonal variation, whereas uniform color distributions (e.g., Setup 8) led to more homogeneous visual compositions.

\begin{table}
	\renewcommand{\arraystretch}{0.5}
	\centering
	\caption{Configurations of  different simulation setups}
	\begin{tabular}{@{}c@{\hspace{0.6em}}c@{\hspace{0.6em}}p{0.35cm}@{}p{0.35cm}@{}p{0.35cm}@{}p{0.35cm}@{}p{0.35cm}@{}p{0.35cm}@{}p{0.35cm}@{}p{0.35cm}@{}p{0.35cm}@{}p{0.35cm}@{}p{0.35cm}@{}p{0.35cm}@{\hspace{0.3em}}|c@{\hspace{0.3em}}c@{\hspace{0.3em}}c@{}}
		\toprule
		\textbf{Setup} & \textbf{Color} & \textbf{1} & \textbf{2} & \textbf{3} & \textbf{4} & \textbf{5} & \textbf{6} & \textbf{7} & \textbf{8} & \textbf{9} & \textbf{10} & \textbf{11} & \textbf{12} & \textbf{N} & \textbf{L} & \textbf{w} \\ \midrule
		\multirow{3}{*}{1}  & C & x & x & x & x & x & x &   &   &   &   &   &   & \multirow{3}{*}{6} & \multirow{3}{*}{1} & \multirow{3}{*}{15} \\ 
		& M & x & x & x & x & x & x &   &   &   &   &   &   &                     &                     &                     \\ 
		& Y & x & x & x & x & x & x &   &   &   &   &   &   &                     &                     &                     \\ \midrule
		\multirow{3}{*}{2}  & C & x & x & x & x & x & x &   x&   x&   x&   &   &   & \multirow{3}{*}{9}& \multirow{3}{*}{1} & \multirow{3}{*}{15} \\ 
		& M & x & x & x & x & x & x &   x&   x&   x&   &   &   &                     &                     &                     \\ 
		& Y & x & x & x & x & x & x &   x&   x&   x&   &   &   &                     &                     &                     \\ \midrule
		\multirow{3}{*}{3}  & C & x & x & x & x & x & x &   x&   x&   x&   x&   x&   x& \multirow{3}{*}{12}& \multirow{3}{*}{1} & \multirow{3}{*}{15} \\ 
		& M & x & x & x & x & x & x &   x&   x&   x&   x&   x&   x&                     &                     &                     \\ 
		& Y & x & x & x & x & x & x &   x&   x&   x&   x&   x&   x&                     &                     &                     \\ \midrule
		\multirow{3}{*}{4}  & C & x & x & x & x & x & x &   &   &   &   &   &   & \multirow{3}{*}{6} & \multirow{3}{*}{3}& \multirow{3}{*}{15} \\ 
		& M & x & x & x & x & x & x &   &   &   &   &   &   &                     &                     &                     \\ 
		& Y & x & x & x & x & x & x &   &   &   &   &   &   &                     &                     &                     \\ \midrule
		\multirow{3}{*}{5}  & C & x & x & x & x & x & x &   &   &   &   &   &   & \multirow{3}{*}{6} & \multirow{3}{*}{5}& \multirow{3}{*}{15} \\ 
		& M & x & x & x & x & x & x &   &   &   &   &   &   &                     &                     &                     \\ 
		& Y & x & x & x & x & x & x &   &   &   &   &   &   &                     &                     &                     \\ \midrule
		\multirow{3}{*}{6}  & C & x & x & x & x & x & x &   &   &   &   &   &   & \multirow{3}{*}{6} & \multirow{3}{*}{1} & \multirow{3}{*}{10}\\ 
		& M & x & x & x & x & x & x &   &   &   &   &   &   &                     &                     &                     \\ 
		& Y & x & x & x & x & x & x &   &   &   &   &   &   &                     &                     &                     \\ \midrule
		\multirow{3}{*}{7}  & C & x & x & x & x & x & x &   &   &   &   &   &   & \multirow{3}{*}{6} & \multirow{3}{*}{1} & \multirow{3}{*}{20}\\ 
		& M & x & x & x & x & x & x &   &   &   &   &   &   &                     &                     &                     \\ 
		& Y & x & x & x & x & x & x &   &   &   &   &   &   &                     &                     &                     \\ \midrule
		\multirow{3}{*}{8}  & C & x & & & x & & &   &   &   &   &   &   & \multirow{3}{*}{6} & \multirow{3}{*}{1} & \multirow{3}{*}{15} \\ 
		& M & & x & & & x & &   &   &   &   &   &   &                     &                     &                     \\ 
		& Y & & & x & & & x &   &   &   &   &   &   &                     &                     &                     \\ \midrule
		\multirow{3}{*}{9}  & C & x & & x& x & & x &   &   &   &   &   &   & \multirow{3}{*}{6} & \multirow{3}{*}{1} & \multirow{3}{*}{15} \\ 
		& M & x & x& & x & x & &   &   &   &   &   &   &                     &                     &                     \\ 
		& Y & & x & x & & x & x &   &   &   &   &   &   &                     &                     &                     \\ \midrule
		\multirow{3}{*}{10} & C & x & & & x & & x &   &   &   &   &   &   & \multirow{3}{*}{6} & \multirow{3}{*}{1} & \multirow{3}{*}{15} \\ 
		& M & & x& & x & x & &   &   &   &   &   &   &                     &                     &                     \\ 
		& Y & & & x & & x & x &   &   &   &   &   &   &                     &                     &                     \\ \midrule
		\multirow{3}{*}{11} & C & x & & x & x & x & x &   &   &   &   &   &   & \multirow{3}{*}{6} & \multirow{3}{*}{1} & \multirow{3}{*}{15} \\ 
		& M & x & x & & x & x & x &   &   &   &   &   &   &                     &                     &                     \\ 
		& Y & & x & x & x & x & x &   &   &   &   &   &   &                     &                     &                     \\ \midrule
		\multirow{3}{*}{12} & C & x & & & x & x & x &   &   &   &   &   &   & \multirow{3}{*}{6} & \multirow{3}{*}{1} & \multirow{3}{*}{15} \\ 
		& M & & x & & x & x & x &   &   &   &   &   &   &                     &                     &                     \\ 
		& Y & & & x & x & x & x &   &   &   &   &   &   &                     &                     &                     \\ \midrule
		\multirow{3}{*}{13} & C & x & & x & x & x & x &   &   &   &   &   &   & \multirow{3}{*}{6} & \multirow{3}{*}{1} & \multirow{3}{*}{15} \\ 
		& M & & x & & x & x & x &   &   &   &   &   &   &                     &                     &                     \\ 
		& Y & & & x & & x & x &   &   &   &   &   &   &                     &                     &                     \\ \bottomrule
	\end{tabular}
	\label{table:setup}
\end{table}

\subsection{Experiments}

\begin{figure*}
    \centering
    \subfloat[]{\label{subfig:moonlightfirst}\includegraphics[trim={3cm 4cm 8cm 4cm},clip,width=0.33\linewidth]{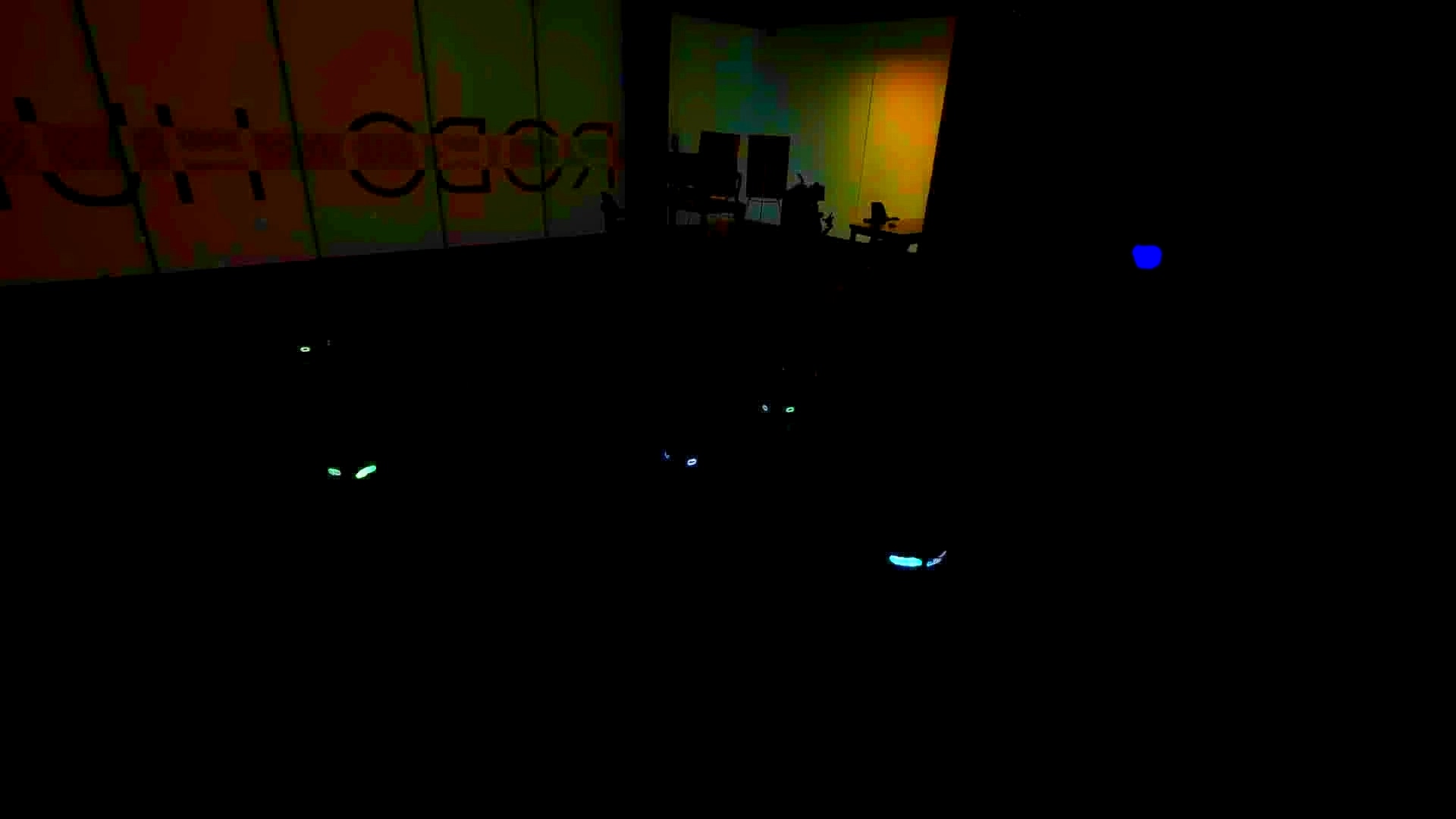}}\hfill
    \subfloat[]{\includegraphics[trim={3cm 4cm 8cm 4cm},clip,width=0.33\linewidth]{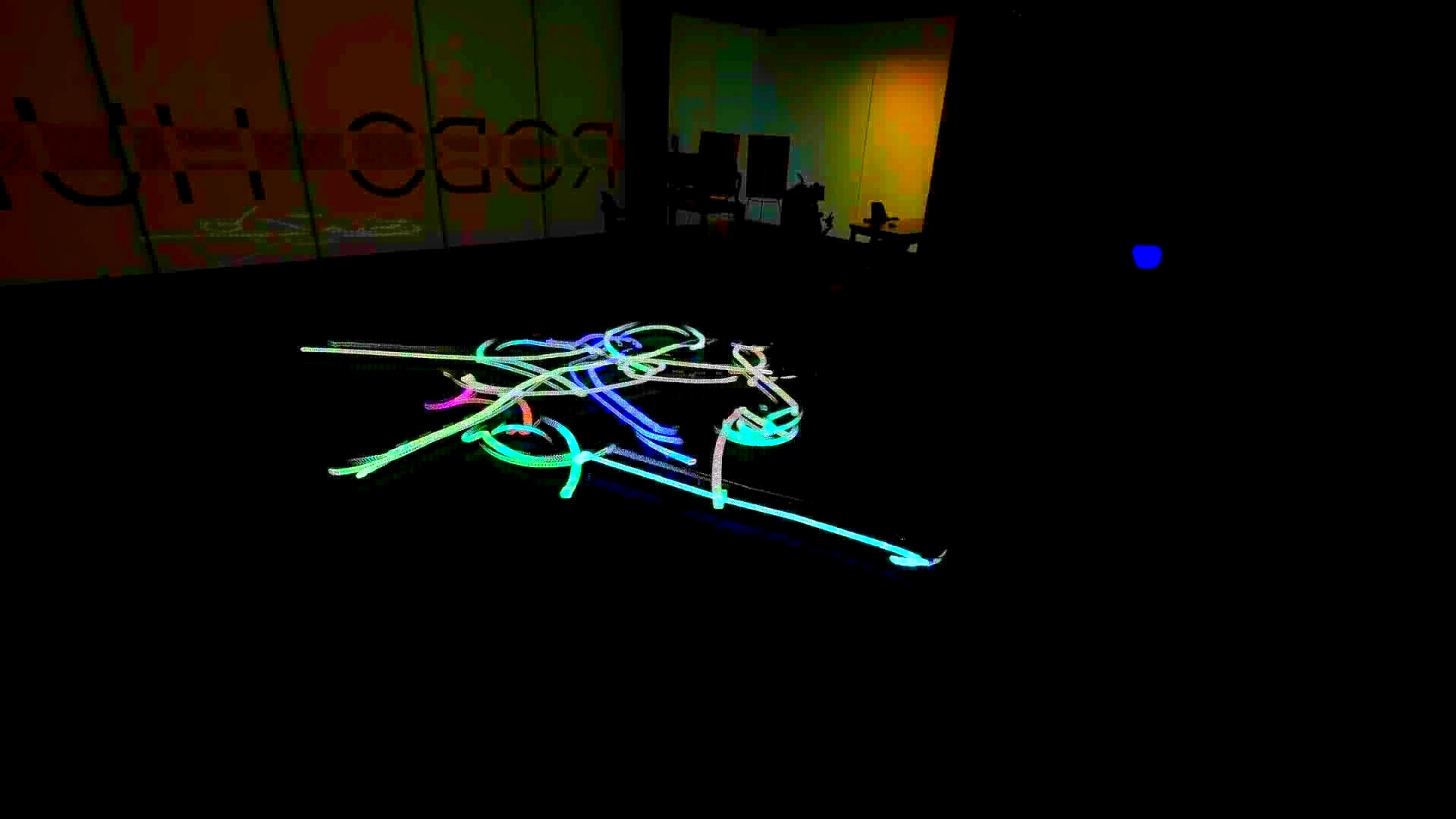}}\hfill
    \subfloat[]{\includegraphics[trim={3cm 4cm 8cm 4cm},clip,width=0.33\linewidth]{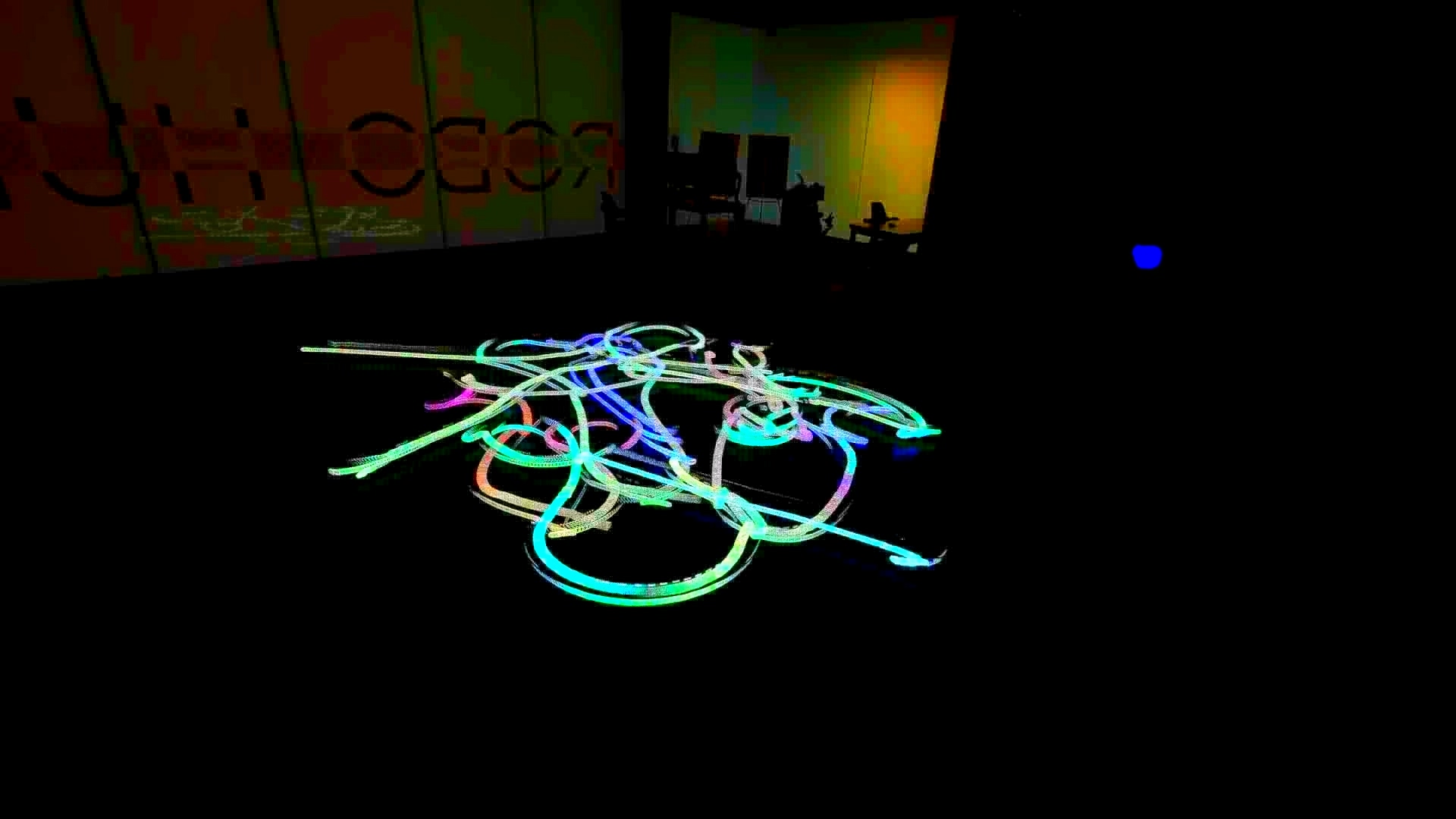}}\\
    \subfloat[]{\includegraphics[trim={3cm 4cm 8cm 4cm},clip,width=0.33\linewidth]{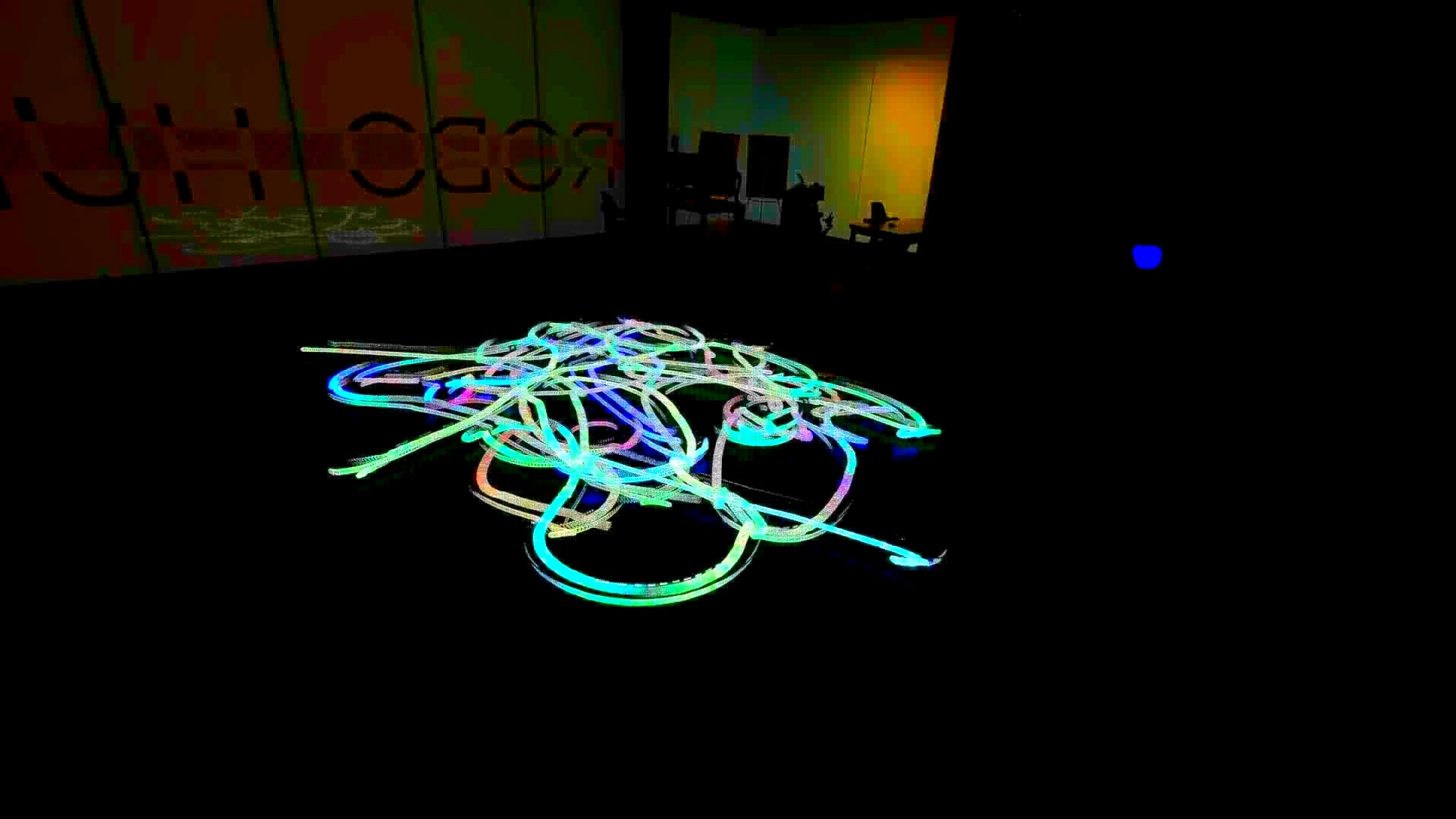}}\hfill
    \subfloat[]{\includegraphics[trim={3cm 4cm 8cm 4cm},clip,width=0.33\linewidth]{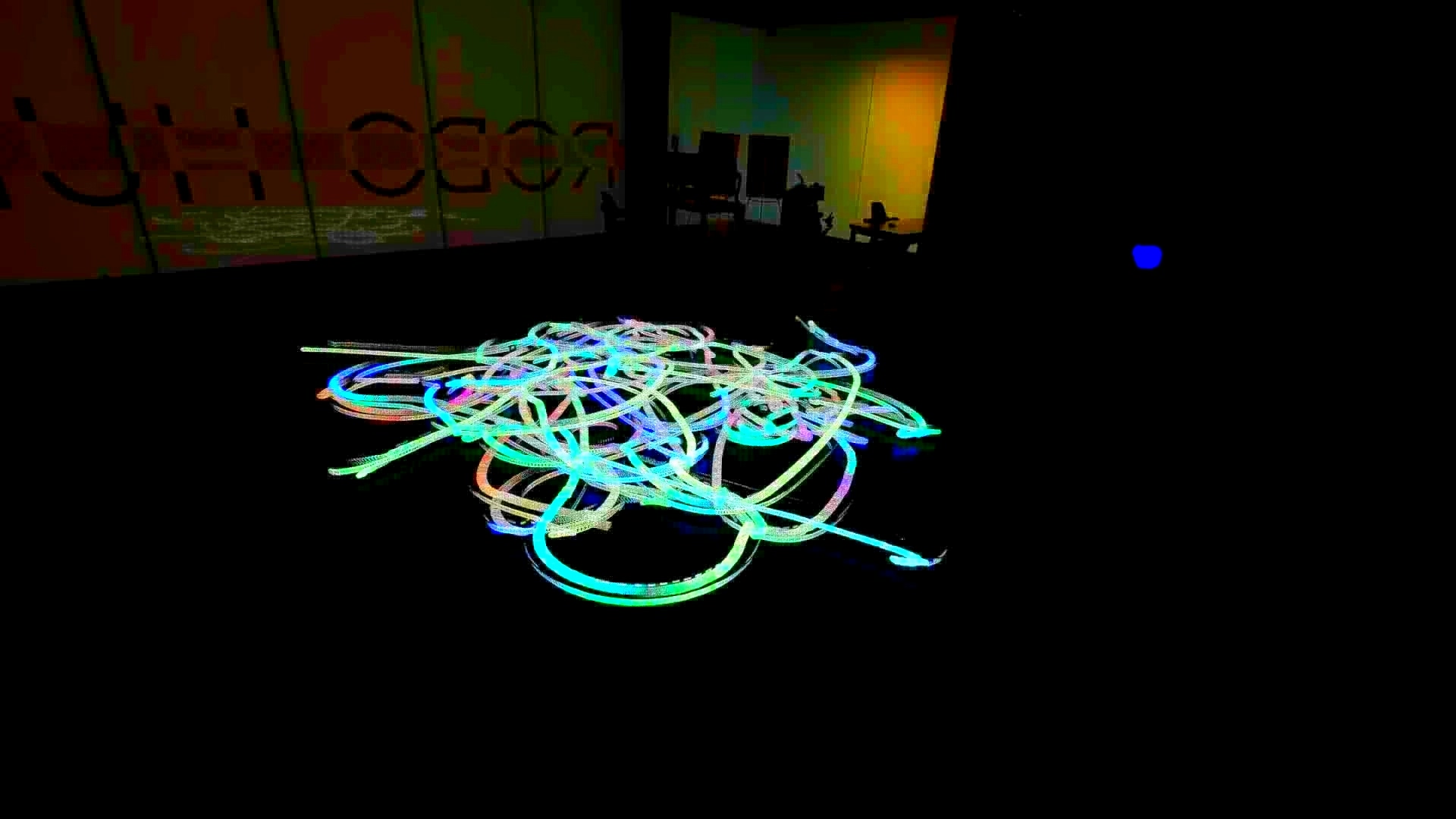}}\hfill
    \subfloat[]{\label{subfig:moonlightlast}\includegraphics[trim={3cm 4cm 8cm 4cm},clip,width=0.33\linewidth]{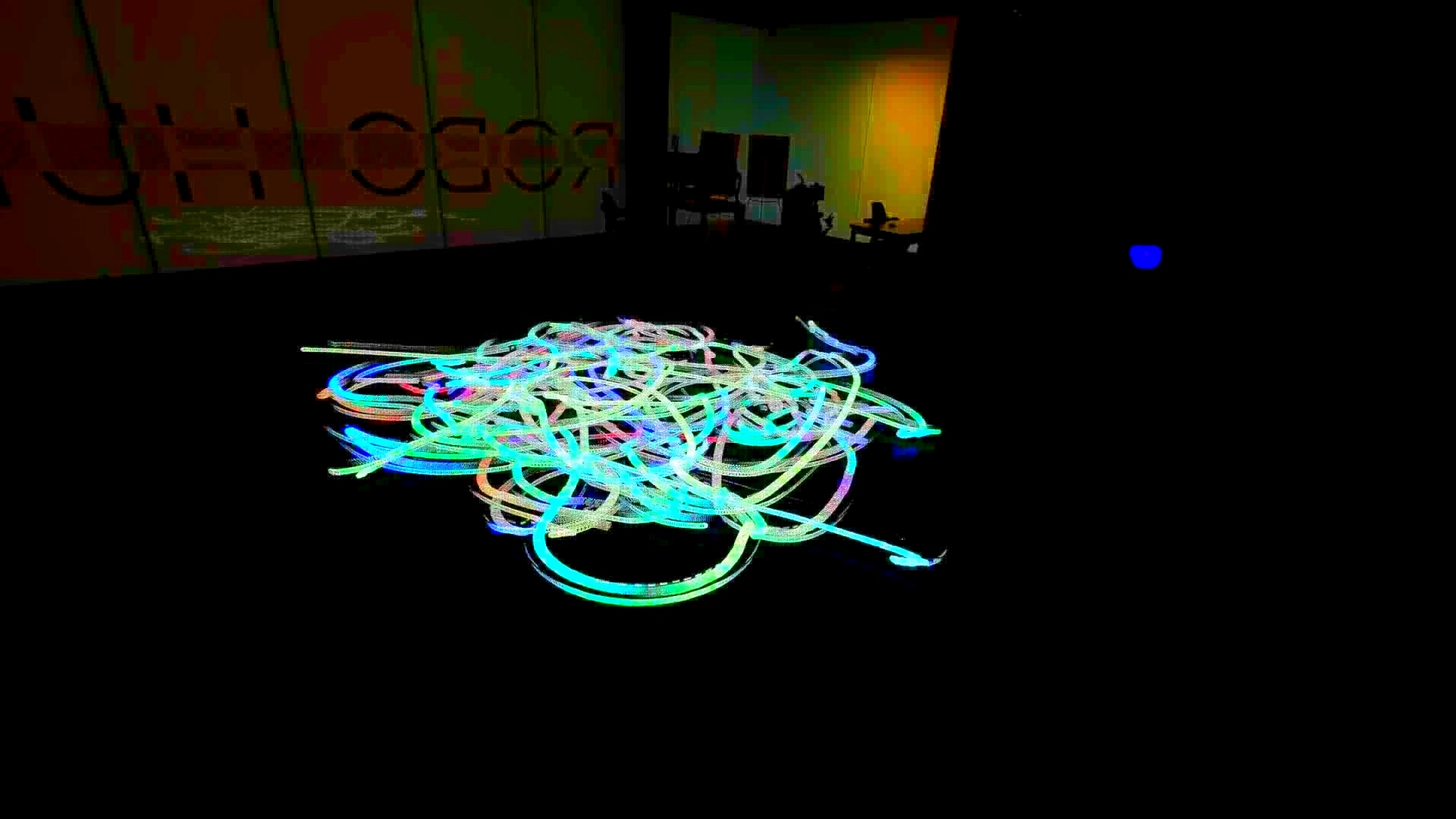}}
    \caption{Six DJI Robomaster EP robots light-painting the Moonlight Sonata. A full video of the experiment with the musical input can be accessed online at \url{https://youtu.be/sLqYoUJDHh4}, a 1-minute extract of which is available as a multimedia attachment.}
    \label{fig:grid_setup2}
\end{figure*}

In our real-robot implementation, we considered the following musical compositions: the Beethoven's \textit{Moonlight Sonata} and the ``lost'' Waltz in A minor, attributed to Frédéric Chopin and unearthed in October 2024. Instead of traditional paint dispensers, each robot in the swarm was equipped with a programmable RGB LED capable of displaying the full color spectrum. By recording the movement of these robots in an environment artificially darkened by reducing the video luminosity, we simulated physical paint using long-exposure light trails that captured each robot’s path and color transition over time. During this process, the physical robots remain visible, allowing users to observe the painting as it unfolds.

The trials with Moonlight Sonata yielded notable results (see snapshots of the experiment in Figures~\ref{subfig:moonlightfirst}--\ref{subfig:moonlightlast}). The robots responded effectively to the composition's tempo and the minor-key harmonic structure. The resulting painting prominently featured blue and purple hues, aligning with our mapping of the melancholic character of the piece. In addition, spatial density observations showed a higher concentration of movement in canvas regions corresponding to minor chords, confirming the proper implementation of our chord-to-location mapping.

Similarly, the trials with Chopin's \textit{Waltz in A Minor} reflected the slower tempo and calm emotional tone of the piece, producing cooler color patterns and smoother trails (Fig.~\ref{fig:lost_waltz}). This demonstrated the ability of the system to translate musical characteristics into visual representations. The consistency between extracted musical features, emotional interpretation, and coordinated robotic movement confirms the potential of the system to generate coherent and expressive visualizations of musical mood and structure.

\section{Conclusions}

In this work, we introduced a novel interactive robotic art system in which a swarm of mobile robots generates expressive visual art based on musical input. By leveraging chord-to-emotion mappings and coverage control, the proposed system creates visualizations that reflect the emotional picture of a musical composition. Through extensive simulations and real-world experiments with light-painting robots, we demonstrated the feasibility and expressive potential of our approach.

Future work will be devoted to designing and running user studies to evaluate the expressiveness of the robotic system when interacting with the creativity of human artists. Moreover, we are interested in characterizing the role of the robot physical embodiment, namely how the presence and movement of real robots shape emotional impact and engagement. We expect that, unlike digital simulations, physical robots offer a more immersive and engaging experience that may stimulate artistic expression and viewer interaction.

\bibliographystyle{bib/IEEEtran}
\bibliography{bib/IEEEabrv,bib/references}

\end{document}